\documentclass{article}
\usepackage{PRIMEarxiv}
\usepackage[T1]{fontenc}
\usepackage[numbers,sort&compress]{natbib}
\usepackage{microtype}
\usepackage{amsmath,amsfonts,bm}

\usepackage{booktabs}
\usepackage{graphicx}
\usepackage{amssymb}
\usepackage{array}
\usepackage{placeins}
\usepackage{flafter}
\usepackage{tabularx}
\usepackage{xcolor}
\usepackage[hidelinks]{hyperref}
\usepackage{url}
\usepackage{rivet_layout}

\AddToHook{env/table/begin}{\setlength{\abovecaptionskip}{0pt}\setlength{\belowcaptionskip}{6pt}}

\newenvironment{tightcontrib}
{\begin{list}{$\bullet$}{%
  \setlength{\leftmargin}{1.15em}%
  \setlength{\labelwidth}{0.65em}%
  \setlength{\labelsep}{0.35em}%
  \setlength{\itemsep}{1pt}%
  \setlength{\parsep}{0pt}%
  \setlength{\topsep}{2pt}%
}}
{\end{list}}

\title{Visual Evidence Under Cross-Examination:\\Evaluating and Controlling Decision-Level\\Evidence Use in Vision-Language Models}
\author{\textbf{Huiyao~Zhang}$^{1,2}$\quad \textbf{Jin~Bai}$^{1,2}$\quad \textbf{Zilong~Su}$^{3}$\quad \textbf{Rui~Guo}$^{1,2}$\\
\textbf{Chaofan~Qin}$^{1,2}$\quad \textbf{Jinze~Lv}$^{1,2}$\quad \textbf{Wenhui~Yu}$^{1,2}$\quad \textbf{Hongfei~Wang}$^{1,2}$\quad \textbf{Ye~Li}$^{1,*}$\\[5pt]
{\normalfont\small $^{1}$ Technology and Engineering Center for Space Utilization, Chinese Academy of Sciences}\\
{\normalfont\small $^{2}$ University of Chinese Academy of Sciences\quad $^{3}$ University of Science and Technology of China}\\[3pt]
{\normalfont\small \href{mailto:zhanghuiyao25@csu.ac.cn}{\texttt{zhanghuiyao25@csu.ac.cn}}\quad \href{mailto:liye@csu.ac.cn}{\texttt{liye@csu.ac.cn}}\quad ($^{*}$ Corresponding author)}}
\date{}
\hypersetup{
  pdftitle={Visual Evidence Under Cross-Examination: Evaluating and Controlling Decision-Level Evidence Use in Vision-Language Models},
  pdfauthor={Huiyao Zhang, Jin Bai, Zilong Su, Rui Guo, Chaofan Qin, Jinze Lv, Wenhui Yu, Hongfei Wang, Ye Li},
  pdfsubject={Candidate-bound visual contribution, CROSS-Bench, and RIVET},
  pdfkeywords={Vision-Language Models, Visual Reasoning, Counterfactual Evaluation, Evidence Attribution, Faithfulness}
}

\begin{document}
\flushbottom
\maketitle

\begin{abstract}

Vision-language models increasingly reason through crops, regions, and tool-produced observations. Yet an observation can influence the answer without benefiting the candidate it supports. We study \emph{candidate-bound visual contribution}: valid evidence should help, invalidating its supporting relation should remove its additional effect, and valid rebinding should redirect that effect to the newly supported candidate. We introduce \textbf{CROSS-Bench}, a benchmark of 28,000 decision problems, with matched invalidation and rebinding tests on a dedicated evaluation subset. Our \textbf{RIVET} interface preserves evidence identity and uncertainty, composes a candidate-conditioned response, and separately controls its strength. Shared-evidence experiments show that task accuracy and evidence ownership can diverge. Under matched capacity and training, RIVET increases normalized effect transfer from 0.512 to 0.651 where clean evidence has a positive effect. The advantage persists on common evaluation examples and across repeated decision-layer fits. With evidence predicted from raw inputs, RIVET improves CROSS-Bench accuracy by an average of 5.70~pp across four frozen backbones, relative to the same models without auxiliary evidence. These results separate the utility of visual evidence from the candidate-specific destination of its effect.

\end{abstract}

\section{Introduction}

Regions, crops, marked images, visual memories, and tool-produced observations expose intermediate visual evidence during vision-language reasoning~\citep{visualcot,vstarbench,setofmark,deepeyes,vlmr3}. A crop can reveal a small attribute that is hard to resolve in the full image, while a region reference identifies the entity relevant to a claim. Grounding supervision further connects intermediate claims to visible entities and relations~\citep{groundedcot,treevgr,regionreasoner}. These interfaces make visual information inspectable, but do not by themselves ensure that its effect reaches the supported candidate. A model may localize the right object and read the right state yet assign the resulting update to an unsupported candidate. Evaluating visual evidence therefore requires examining both its content and the candidate-level destination of its effect.

\begin{figure}[!t]
    \centering
    \includegraphics[width=\linewidth]{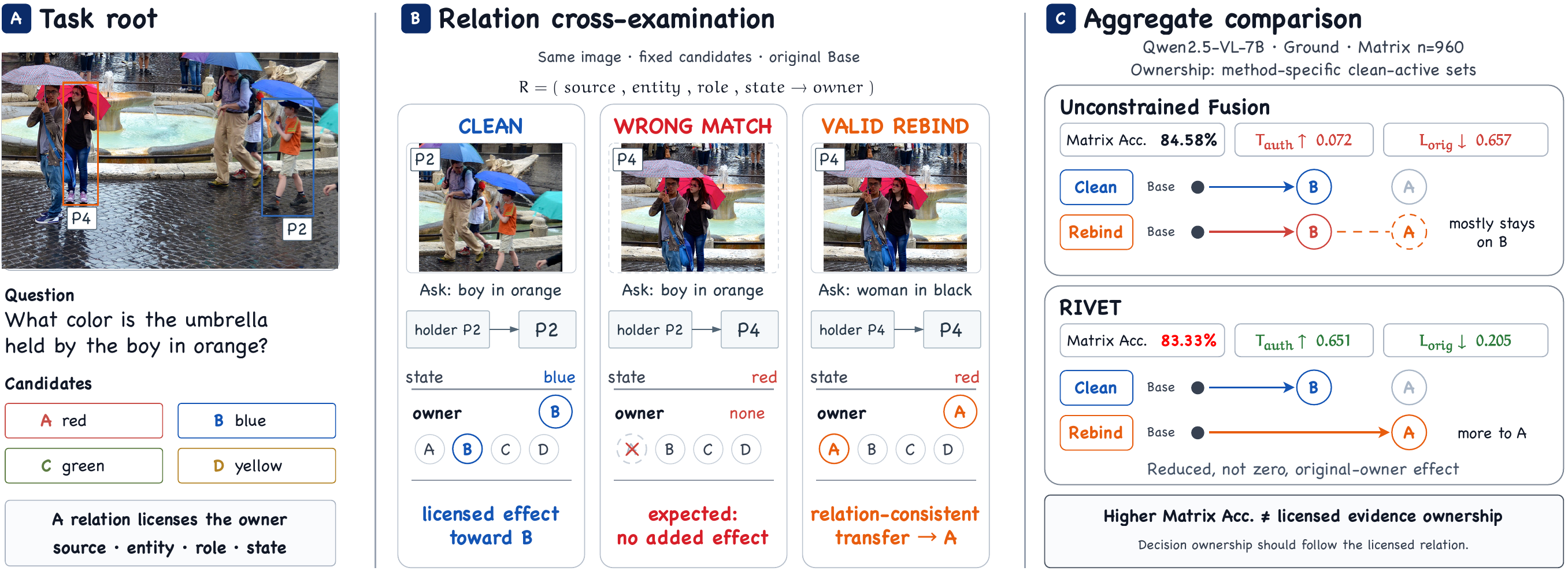}
    \caption{\textbf{Candidate-bound visual contribution.}
    (A) Two umbrella--holder relations in one image;
    (B) clean, wrong-match, and valid-rebind pairs with the same image and original Base;
    (C) aggregate utility--ownership comparison on the 960-root matrix (Qwen2.5-VL-7B, predicted evidence).
    Ownership uses method-specific clean-active sets (612 and 521 roots, respectively).
    Arrows denote schematic effect destinations, not magnitudes.}
    \label{fig:ownership-thesis}
\end{figure}

Consider the two umbrellas in Fig.~\ref{fig:ownership-thesis}. The boy in orange holds a blue umbrella, while the woman in black holds a red one. An observation of the boy's umbrella can add support for blue when the question concerns him. A view of the woman's umbrella may also yield a correct red reading, yet that reading does not support red for the boy's umbrella. When the request changes to the woman, the same red observation becomes relevant and should increase support for red. Across the latter two conditions, the observation and its state remain unchanged; the requested holder determines its relevance.

Existing work improves visual access, grounds intermediate claims, and diagnoses whether perceived information enters the decision~\citep{seeingnotbelieving,cogflow,composefuse}. Counterfactual objectives encourage visual dependence and evidence--answer consistency~\citep{papo,defacto}. Self-Critical Reasoning regularizes correct and competing answers' visual sensitivities~\citep{selfcriticalvqa}, while receiver-dependent evidence integration models how external evidence shifts an existing answer distribution~\citep{evidenceintegration}. These studies make candidate-level influences explicit, alongside work on visual dependence and the faithfulness of explanations~\citep{faithfulcot,faithlm}. A complementary question is \emph{which candidate receives the observation's effect as its supporting relation changes?} Correctness may follow from priors or other evidence, while corruption sensitivity reveals input dependence without identifying the effect's destination. We call this property \emph{decision ownership}: whether an established evidence effect follows the answer candidate supported by the current visual relation.

We formalize the broader criterion as \emph{candidate-bound visual contribution}. \textbf{Utility} measures the benefit of valid evidence; \textbf{specificity} tests whether invalidating its supporting relation removes the additional effect; and \textbf{ownership} asks whether an established effect follows a valid change of owner. Suppressing every observation can pass invalidation without using evidence. Conversely, rewarding a fixed candidate can improve accuracy without following the visual relation. Valid rebinding supplies the positive requirement missing from either behavior: the established effect should redirect to the newly supported candidate. The criterion evaluates changes in the candidate posterior rather than only whether the top answer changes. When the original input already favors one answer, valid evidence can still add support, and a wrongly directed update can remain hidden behind the same final prediction.

\textbf{CROSS-Bench} operationalizes this criterion through matched clean, invalid, and rebinding conditions on a dedicated evaluation subset of its 28,000 decision roots. The benchmark covers visual attributes, counts, spatial relations, and text/chart/graph states. Effects are measured relative to Base, which sees the original image, question, and answer candidates but does not use auxiliary observations. The same predicted observations can be supplied to different decision interfaces, so differences in their candidate-level effects need not arise from different visual readings. Full-suite task accuracy provides a complementary view of the predictive utility of the evidence.

The criterion also motivates separating \emph{which candidates the evidence supports} from \emph{how strongly it should act}. An observation's relevance depends on its source and semantic role, which a confidence score alone does not specify. These relations must remain available when composing multiple observations for each candidate. \textbf{RIVET} preserves evidence identity, bindings, candidate alignment, and uncertainty across specialized visual readers, composes evidence around candidate queries, and then controls response strength with a scalar. Candidate-conditioned composition gives each alternative access to the full evidence set, so the response can depend on how observations support that alternative jointly. The scalar controller operates on this composed response, retaining a distinction between the evidence's candidate alignment and the size of the resulting update. Uncertainty accompanies the observations throughout composition, so an uncertain reading need not become complete support for a candidate.

\noindent Our contributions are threefold:
\begin{tightcontrib}
\item We introduce \emph{candidate-bound visual contribution} and \textbf{CROSS-Bench}, a 28,000-root benchmark with paired tests on a dedicated subset that evaluate whether visual evidence helps when valid, loses its additional effect when invalidated, and transfers that effect under valid rebinding.
\item We develop \textbf{RIVET}, a shared evidence-to-decision interface that preserves identity, candidate alignment, and uncertainty across specialized visual readers, composes evidence through candidate-conditioned representations, and separately controls response strength with a scalar posterior controller.
\item We show that task accuracy and evidence ownership can diverge under matched evidence. Relative to a capacity-matched fusion control under matched training, RIVET increases effect transfer, normalized to the clean effect, from 0.512 to 0.651 on each method's clean-active roots. With evidence predicted from raw inputs, it also improves task accuracy over Base by an average of 5.70~pp across four frozen backbones.
\end{tightcontrib}

\section{Related Work}

\noindent\textbf{Visual evidence acquisition and grounding.}
Visual reasoning interfaces expose local content through crops, marks, and region references~\citep{visualcot,setofmark}. Visual CoT pairs intermediate region annotations with a multi-turn pipeline, while Set-of-Mark assigns regions visual identifiers that can be referenced in language. Beyond fixed observations, V$^\star$ uses language-guided search to recover details in high-resolution images~\citep{vstarbench}. Tool-using models acquire additional views during reasoning~\citep{deepeyes,vlmr3}: DeepEyes learns active visual inspection, and VLM-R$^3$ learns when and where to acquire additional views and incorporate them into an interleaved reasoning chain. Localization supervision and grounded chains of thought connect intermediate claims to visible objects and relations. Grounded Chain-of-Thought studies stepwise localization and answer--grounding consistency~\citep{groundedcot}; TreeVGR jointly supervises localization and reasoning~\citep{treevgr}; and RegionReasoner requires explicit region references and encourages global--local consistency~\citep{regionreasoner}. These approaches improve visual access and the traceability of reasoning.

\noindent\textbf{Evidence utilization and counterfactual dependence.}
Multimodal benchmarks assess visual necessity and fine-grained perception~\citep{mmstar,mmerealworld}. Diagnostic studies examine perception--reasoning gaps~\citep{seeingnotbelieving,cogflow,composefuse,spdfaith}. Seeing but Not Believing distinguishes attention to relevant content from its use in a decision; CogFlow studies the integration of perceived cues; and Compose and Fuse examines bottlenecks in task composition and cross-modal fusion. Counterfactual objectives probe visual dependence~\citep{papo}, including evidence--answer consistency under image masking~\citep{defacto}. Such interventions examine whether predictions change when relevant visual information is removed or corrupted. Faithfulness studies test explanations against decisions~\citep{faithfulcot,faithlm}, and Self-Critical Reasoning regularizes correct and competing answers' visual gradient sensitivities~\citep{selfcriticalvqa}. Visual evidence selection evaluates output-distribution utility~\citep{utilityvisualevidence}, while receiver-dependent integration models candidate-specific evidence relative to a prior~\citep{evidenceintegration}. Detecting an influence under a given support relation does not by itself establish how that influence changes when the relation becomes invalid or validly supports a different candidate.

\noindent\textbf{Visual binding and compositionality.}
Attribute and relation benchmarks test whether models distinguish compositions built from the same objects and concepts. ARO examines attribute assignment, relations, and word order~\citep{aro}, while Winoground requires distinguishing image--caption pairs whose captions contain the same words~\citep{winoground}. Careful negative construction is needed to avoid linguistic shortcuts; SugarCrepe addresses this issue through fluent hard negatives that reduce linguistic artifacts in negative captions~\citep{sugarcrepe}. Entity-grounded composition and reusable visual programs structure visual reasoning~\citep{covlm,genome}. CoVLM interleaves language generation with visual detection, retaining explicit visual entities and relations as text is generated. GENOME constructs executable visual modules and reuses them across tasks, making the reasoning procedure available as a sequence of operations rather than a single holistic prediction. Mechanistic analyses examine internal feature binding~\citep{visualsymbolicmechanisms}, including how visual features are associated with objects and how failures in these associations relate to binding errors.

\section{Candidate-Bound Visual Contribution}
\label{sec:formulation}

We compare visual evidence effects over a finite, ordered candidate set $\mathcal C_i$. Each decision problem, or root, $r_i=(I_i,q_i,\mathcal C_i,y_i)$ contains an image, request, candidates, and clean answer. The Base posterior $\mathbf p_i^0$ uses the same image, request, candidates, and scoring procedure, without auxiliary observations. For each evidence condition $\iota$, the evaluator records the support relation
\begin{equation}
\mathcal R_i^\iota:
\left(
\mathrm{src},
\mathrm{ent},
\mathrm{role},
\mathrm{state}
\right)_i^\iota
\mapsto o_i^\iota,
\label{eq:contribution-relation}
\end{equation}
where $o_i^\iota\in\mathcal C_i\cup\{\varnothing\}$ is the candidate supported by this relation, with $\varnothing$ denoting no valid owner. Each trial supplies an observable evidence--request pair $u_i^\iota$. The request specifies the relevant source, entity, and roles without revealing the answer. Ground-truth states, owners, and intervention labels remain evaluator-side.

The same candidate interface answers each condition's supplied request. Clean trials address $q_i$; valid rebinding addresses $q_i'$ with matching evidence. The primary rebinding test retains the original root, candidate order, and stored $\mathbf p_i^0$ as references. It measures the new owner's effect relative to the original Base when the request and supporting relation change together. A complementary test fixes $q_i'$, recomputes Base and evidence-conditioned responses, and varies evidence relevance. This measures within-request selectivity, complementing the cross-request transfer test. Both tests apply to any system exposing matched candidate posteriors, without requiring RIVET's typed representation.

\section{CROSS-Bench: Evaluating Candidate-Bound Evidence Use}
\label{sec:benchmark}

CROSS-Bench operationalizes this criterion through paired evidence interventions. It contains 28,000 decision roots constructed from public annotations, including Visual Genome, GQA, and TextOCR~\citep{visualgenome,gqa,textocr}, and deterministic renderings, with a frozen test set of 5,600 roots. Each root pairs an image and question with $N_i=|\mathcal C_i|\in\{2,3,4,5\}$ candidates and a source--entity--role--state relation. Accuracy uses the clean question; primary relation probes vary the evidence relation against the same Base reference (Fig.~\ref{fig:ownership-thesis}).

A prediction-blind subset of 960 test roots supports complete relation interventions, with 707 admitting valid rebinding. This subset diagnoses evidence use; all 5,600 roots measure full-suite accuracy. Train/dev/test partitions are source-cluster-disjoint. Across 1,200 reviewed roots, both reviewers confirm unique valid targets on 98.8\% and answer neutrality on 99.3\%; they also assess alternative-owner validity.

\begin{figure}[!tb]
\centering
\includegraphics[width=\linewidth]{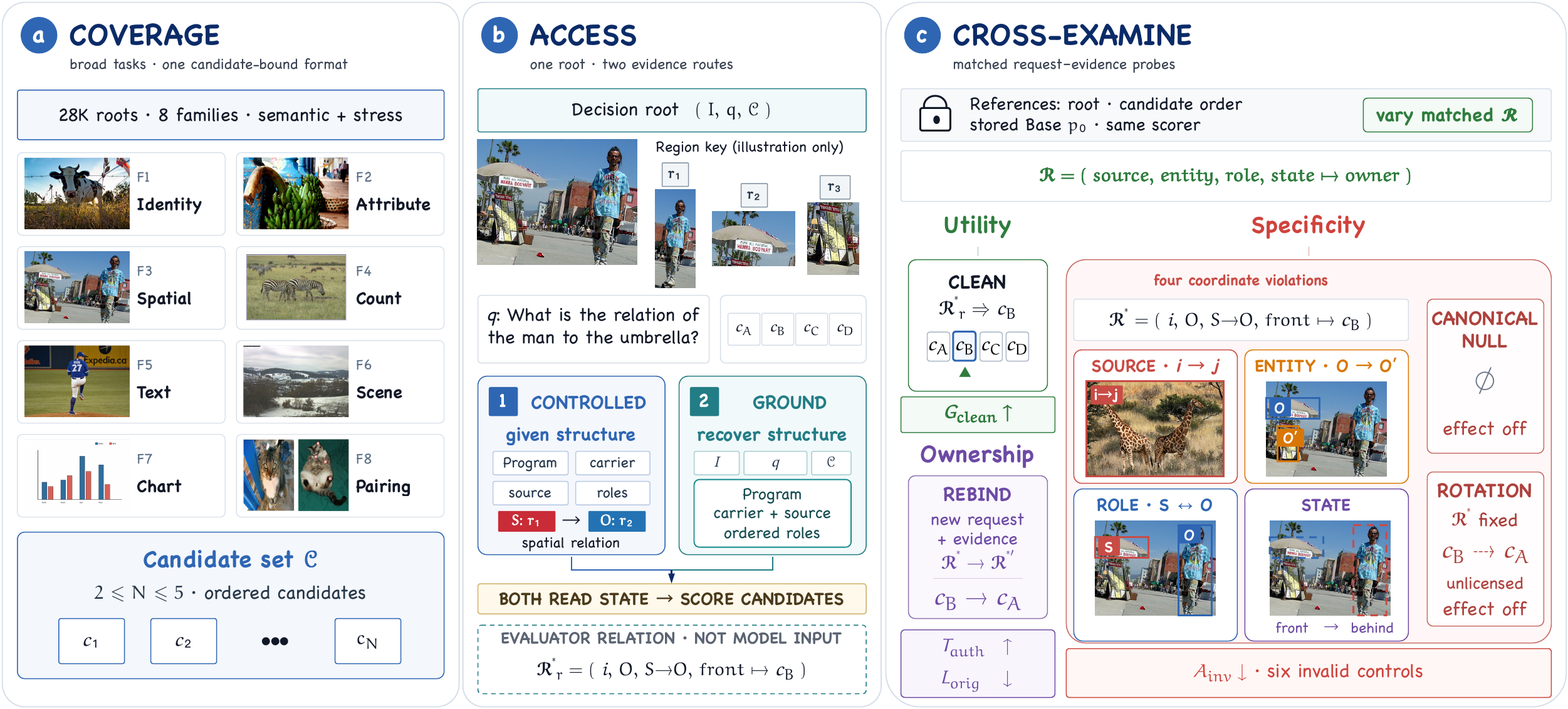}
\caption{\textbf{CROSS-Bench evaluation protocol.} Controlled supplies evidence-access structure and Ground predicts it; both estimate visual state. Clean, invalid, and rebind conditions test utility, specificity, and ownership over 2--5 candidates. ``Effect off'' denotes the expected absence of additional contribution.}
\label{fig:benchmark-overview}
\end{figure}

\noindent\textbf{Two evidence-access routes.}
\textbf{Controlled} supplies the required-fact specification, source identity, and ordered roles; \textbf{Ground} predicts them from raw inputs. Both infer typed state without owner or intervention labels. Crops in Fig.~\ref{fig:benchmark-overview}(b) illustrate the carriers; they are not extra Ground inputs. Ground predicts access before replay. Observation replacements restart acquisition; role changes reuse observations but rerun reading and subsequent stages. Typed-state replay replaces state at fixed bindings before the shared decision suffix.

For each matrix root, \textbf{clean} preserves the relation and \textbf{canonical null} removes auxiliary contribution. Wrong-source, wrong-entity, wrong-role, and wrong-state invalidate individual coordinates; unauthorized \textbf{candidate rotation} breaks evidence--candidate association without reordering candidates. These invalid conditions should remove the additional effect.

\textbf{Valid rebinding} changes the request and supporting relation together, so the established effect should follow the newly supported candidate. In the umbrella example, asking about the woman in black makes the same red observation relevant. Requests identify the holder and attribute without stating the color. A companion control fixes the new request and varies its evidence.

We summarize utility by clean gain $G_{\mathrm{clean}}$, specificity by invalid activity $A_{\mathrm{inv}}$, and ownership by authorized transfer $T_{\mathrm{auth}}$ and original-owner leakage $L_{\mathrm{orig}}$. Let $\mathbf p_i^\iota$ denote the posterior under condition $\iota$, let $\bar p_{i,c}^{\iota}=\max(p_{i,c}^{\iota},\epsilon)$ with numerical floor $\epsilon=10^{-8}$, and let $\mathcal I_{\mathrm{inv}}$ contain canonical null, single-coordinate relation violations, and unauthorized rotation:
\begin{equation}
\begin{aligned}
G_{\mathrm{clean}}
&=
\mathbb E_i
\left[
\log
\frac{\bar p_{i,y_i}^{\mathrm{clean}}}
     {\bar p_{i,y_i}^{0}}
\right],\\
A_{\mathrm{inv}}
&=
\mathbb E_i
\left[
\max_{\iota\in\mathcal I_{\mathrm{inv}}}
\operatorname{TV}
\left(
\mathbf p_i^\iota,\mathbf p_i^0
\right)
\right].
\end{aligned}
\label{eq:cross-clean-invalid}
\end{equation}

$G_{\mathrm{clean}}$ measures the correct candidate's log-score change, including improvements that leave the argmax unchanged. With $\operatorname{TV}(\mathbf p,\mathbf q)=\tfrac12\|\mathbf p-\mathbf q\|_1$, $A_{\mathrm{inv}}$ captures each root's largest posterior movement across invalid relations, rather than averaging across invalid conditions.

Ownership tests an effect that clean evidence has already established. We therefore evaluate it on scoreable rebinding-eligible roots where clean evidence increases the original owner's probability; this clean-active set is method-specific. Under clean evidence, $o_i=y_i$; valid rebinding introduces another valid relation supporting a new owner $n_i\neq o_i$. With $[x]_+=\max(x,0)$, define $\Delta_{i,c}^{\iota}=p_{i,c}^{\iota}-p_{i,c}^{0}$, $\mathcal D^+=\{i:\Delta_{i,o_i}^{\mathrm{clean}}>\delta_\Delta\}$ for clean-effect threshold $\delta_\Delta=0.010$ (one percentage point), and $b_i=[\Delta_{i,o_i}^{\mathrm{clean}}]_++\epsilon$:
\begin{equation}
\begin{aligned}
T_{\mathrm{auth}}
&=
\mathbb E_{i\in\mathcal D^+}
\left[
\min\!\left(
1,\,
\frac{[\Delta_{i,n_i}^{\mathrm{rebind}}]_+}{b_i}
\right)
\right],\\
L_{\mathrm{orig}}
&=
\mathbb E_{i\in\mathcal D^+}
\left[
\frac{[\Delta_{i,o_i}^{\mathrm{rebind}}]_+}{b_i}
\right].
\end{aligned}
\label{eq:cross-ownership}
\end{equation}

Relative to the original Base, $T_{\mathrm{auth}}$ measures the positive new-owner increment and $L_{\mathrm{orig}}$ the positive original-owner residual, both normalized by the clean effect. These quantities are not success rates. Unnormalized candidate changes provide the complementary effect scale. Transfer is capped; leakage remains uncapped to expose amplification. Fixed-request controls instead use each request's own Base to measure evidence selectivity.

\begin{figure}[!tb]
\centering
\includegraphics[width=\linewidth]{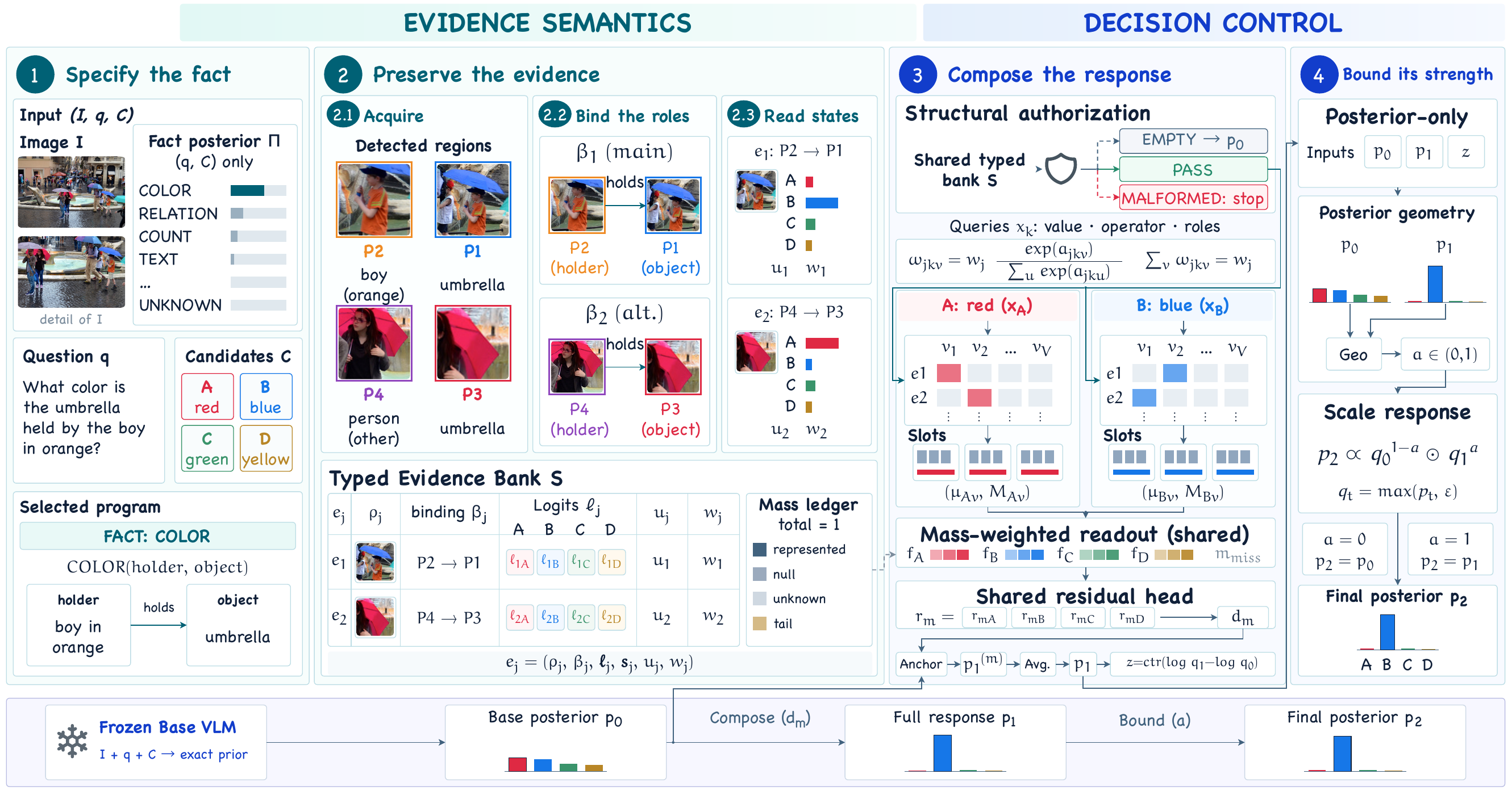}
\caption{\textbf{Overview of RIVET.}
The color example retains alternative bindings in one clean execution. \textsc{Specify} plans visual requirements from $(q,\mathcal C)$; \textsc{Preserve} acquires, binds, and reads observations into a shared typed evidence bank. \textsc{Compose} conditions atom-to-slot assignment on each candidate before mass-weighted aggregation; \textsc{Bound} scales the resulting posterior response. Posterior plots share a unit scale; bank entries retain logits. Here $\mathbf q_t=\max(\mathbf p_t,\epsilon)$ for $t\in\{0,1\}$.}
\label{fig:method-overview}
\end{figure}

\section{RIVET: A Candidate-Bound Evidence-to-Decision Interface}
\label{sec:method}

RIVET preserves evidence identity, bindings, and candidate meaning through composition, then separately scales the response (Fig.~\ref{fig:method-overview}). \textsc{Specify} identifies required facts; \textsc{Preserve} binds and reads observations; \textsc{Compose} organizes evidence around each candidate; and \textsc{Bound} controls response strength.

For ordinary task inference with candidate set $\mathcal C$, we omit the root index and use subscripts $0,1,2$ for decision endpoints:
\begin{equation}
\begin{aligned}
\boldsymbol\Pi
&=
\operatorname{Specify}(q,\mathcal C),
&\qquad
\mathcal S
&=
\operatorname{Preserve}(I,\boldsymbol\Pi,\mathcal C),\\
\mathbf p_1
&=
\operatorname{Compose}_{\theta}(\mathbf p_0,\mathcal S,\mathcal C),
&\qquad
\mathbf p_2
&=
\operatorname{Bound}(\mathbf p_0,\mathbf p_1,\mathbf z).
\end{aligned}
\label{eq:rivet-overview}
\end{equation}

The three endpoints are anchored to the same Base posterior, with $D_0=\mathbf p_0$. $D_1=\mathbf p_1$ incorporates the full evidence response, whose Base-relative log change defines $\mathbf z$ in Eq.~\ref{eq:rivet-compose}; $D_2=\mathbf p_2$ controls its strength.

Primary relation probes supply $u_i^\iota$ (Sec.~\ref{sec:formulation}) at the stage receiving the changed evidence or request, then rerun subsequent steps against the same Base.

\subsection{Specify and Preserve}

\noindent\textbf{Required-fact planning.}
The question and complete candidate set determine which visual distinctions matter: color alternatives require an attribute, whereas spatial alternatives require an ordered relation. \textsc{Specify} predicts these requirements from $(q,\mathcal C)$ without image or Base access. Its output $\boldsymbol\Pi$ retains uncertainty over registered requirements, specifying what to observe rather than a preferred answer.

\textsc{Preserve} executes the active requirements. Acquisition obtains observations; a shared Binder retains weighted assignments to semantic roles. Specialized Readers estimate attributes, counts, spatial relations (direction, relative depth, and proximity), or text/chart/graph states for these bindings. Their distinct state vocabularies map into one candidate-aligned interface for the shared Composer. One observation may support multiple requirements with separate identities and weights: shared pixels need not imply the same semantic role.

\begin{samepage}
\noindent\textbf{Candidate-bound support.}
Reading ``red'' is insufficient without knowing whose umbrella was observed. Each atom joins provenance $\rho_j$ and ordered binding $\beta_j$ with candidate-aligned Reader logits $\boldsymbol\ell_j$, their centered form $\mathbf s_j$, uncertainty $u_j$, and represented mass $w_j$. Here $u_j$ summarizes evidence uncertainty, while $w_j$ is the share of upstream hypothesis probability represented by this atom. With $N=|\mathcal C|$ and $\operatorname{ctr}(\mathbf v)=\mathbf v-N^{-1}(\mathbf 1^\top\mathbf v)\mathbf 1$,
\begin{equation}
\begin{aligned}
e_j
&=
\left(
\rho_j,\beta_j,\boldsymbol\ell_j,\mathbf s_j,u_j,w_j
\right),
&
\mathbf s_j
&=
\operatorname{ctr}(\boldsymbol\ell_j),
\qquad
\mathbf 1^\top\mathbf s_j=0.
\end{aligned}
\label{eq:rivet-support-projection}
\end{equation}
\end{samepage}

Binding identifies whose state is read; a registered deterministic map aligns the Reader state to candidate values in $\boldsymbol\ell_j$. Subtracting the candidate mean yields centered relative evidence $\mathbf s_j$. Unrepresented mass remains in null, unknown, or tail channels, so surviving observations are not promoted to complete support by renormalization.

\subsection{Compose: One Full-Evidence Response}

\textsc{Compose} turns the aligned observations into a joint candidate response. A deterministic structural check verifies the operator, candidate order, positive mass, and provenance, yielding $\mathcal S_{\mathrm{valid}}$. Reading or binding errors with valid schemas can pass these checks. All authorized atoms participate; an empty evidence set leaves Base unchanged. The full response and its centered log-posterior change are
\begin{equation}
\begin{aligned}
\mathbf p_1
&=
\operatorname{Compose}_{\theta}
\left(
\mathbf p_0,\mathcal S_{\mathrm{valid}},\mathcal C
\right),\\
\mathbf q_t
&=
\max(\mathbf p_t,\epsilon),
\qquad t\in\{0,1\},\\
\mathbf z
&=
\operatorname{ctr}
\left(
\log\mathbf q_1-\log\mathbf q_0
\right).
\end{aligned}
\label{eq:rivet-compose}
\end{equation}

Candidate query $\mathbf x_k$ encodes its value, operator, and roles. With atom embeddings and aligned Reader logits, it determines assignment scores $a_{jkv}$ from atom $j$ to slot $v$:
\[
\omega_{jkv}=w_j\frac{\exp a_{jkv}}{\sum_u\exp a_{jku}},
\qquad \sum_v\omega_{jkv}=w_j.
\]
The assignments let each candidate organize the same atoms while preserving every atom's mass $w_j$. Mass-weighted cross-attention yields candidate representations $\mathbf f_k$; $m_{\mathrm{miss}}$ denotes unresolved-mass summaries, with null, unknown, and tail channels kept separate. A shared head in each of five separately fitted members produces a Base-free residual $\mathbf r_m$ from the readouts and candidate embeddings. Centering, elementwise $\tanh$, and a learned positive scale yield $\mathbf d_m$. Adding $\mathbf d_m$ to Base log probabilities and applying softmax gives each member's posterior; their average is $\mathbf p_1$.

\subsection{Bound: Strength-Only Control}

Once \textsc{Compose} forms the response, \textsc{Bound} controls its strength. $\operatorname{Geo}_N$ collects permutation-invariant summaries of posterior uncertainty, disagreement, and the centered response. These determine one scalar $a$, not a candidate-specific correction. When structurally valid support exists,
\begin{equation}
\begin{aligned}
a
&=
\sigma\!\left[
A_\eta\!\left(
\operatorname{Geo}_N(\mathbf p_0,\mathbf p_1,\mathbf z)
\right)
\right],\\
\mathbf p_2
&=
\begin{cases}
\mathbf p_0,
& a=0,\\
\operatorname{Normalize}
\!\left(
\mathbf q_0^{\,1-a}\odot\mathbf q_1^{\,a}
\right),
& 0<a<1,\\
\mathbf p_1,
& a=1.
\end{cases}
\end{aligned}
\label{eq:rivet-bound}
\end{equation}

The learned branch yields $0<a<1$; $a=0$ and $a=1$ define exact Base and full-response control endpoints. For any candidates $c,k$, the interior branch satisfies
\begin{equation}
\log\frac{p_{2,c}}{p_{2,k}}
-
\log\frac{q_{0,c}}{q_{0,k}}
=
a
\left[
\log\frac{q_{1,c}}{q_{1,k}}
-
\log\frac{q_{0,c}}{q_{0,k}}
\right].
\label{eq:rivet-logodds}
\end{equation}

Equation~(\ref{eq:rivet-logodds}) scales every evidence-induced pairwise log-odds change by the same nonnegative scalar, preserving its sign. \textsc{Bound} attenuates the existing response; candidate-wise control would additionally allow its effect to be redirected across candidates.

\noindent\textbf{Training.}
The \textsc{Specify} planner and Binder use separate program and role supervision. Composer learns task likelihood and calibration on CROSS-Bench train. \textsc{Bound} minimizes NLL on fixed posterior pairs from Composer members fitted without gradient updates on those roots; Bound fitting and checkpoint selection use disjoint roots. Acquisition and Readers remain frozen; projection and authorization are deterministic. Owner maps and evaluation-condition labels supervise no component.

\begin{table}[!tb]
\caption{\textbf{CROSS-Bench results on the 960-root intervention matrix.}
(a) Released-system diagnosis; posterior metrics use each checkpoint's no-auxiliary reference.
(b) Shared predicted evidence and Qwen2.5-VL-7B Base.
Clean-active counts give scoreable positive-effect roots out of 707 rebinding-eligible roots; ownership uses these method-specific sets (``--'': undefined).}
\label{tab:cross-main}
\centering
\footnotesize
\renewcommand{\arraystretch}{1.04}
\setlength{\tabcolsep}{3.5pt}
\begin{tabularx}{\linewidth}{
@{}
>{\raggedright\arraybackslash}X
>{\centering\arraybackslash}p{1.35cm}
*{2}{>{\centering\arraybackslash}p{1.15cm}}
>{\centering\arraybackslash}p{1.75cm}
>{\centering\arraybackslash}p{2.40cm}
@{}}
\toprule
Method
& Acc.\ (\%)$\uparrow$
& $G_{\mathrm{clean}}\uparrow$
& $A_{\mathrm{inv}}\downarrow$
& \mbox{Clean-active}
& $T_{\mathrm{auth}}\uparrow\,/\,L_{\mathrm{orig}}\downarrow$ \\
\midrule
\multicolumn{6}{@{}l}{\textbf{(a) Released-system diagnosis}} \\
Qwen2.5-VL-7B (Base)
& 64.90 & 0.000 & 0.000 & 0/707 & -- \\
VLM-R$^3$-7B
& 72.50 & 0.153 & 0.162 & 355/707 & 0.413 / 0.274 \\
Pixel-Reasoner-7B
& 71.04 & 0.128 & 0.143 & 332/707 & 0.427 / 0.253 \\
DeepEyes-7B
& 75.10 & 0.184 & 0.171 & 386/707 & 0.462 / 0.283 \\
TreeVGR-7B
& 73.96 & 0.161 & 0.187 & 367/707 & 0.497 / 0.308 \\
\midrule
\multicolumn{6}{@{}l}{\textbf{(b) Shared-evidence interface comparison}} \\
Direct Evidence Prompt
& 70.73 & -0.241 & 0.310 & 248/707 & 0.107 / 0.899 \\
Pooled Support Fusion
& 80.94 & 0.218 & 0.477 & 574/707 & 0.190 / 0.362 \\
Calibrated Aligned Sum
& 79.79 & 0.221 & 0.000 & 487/707 & 0.603 / 0.237 \\
Unconstrained Fusion
& \textbf{84.58} & \textbf{0.296} & 0.346 & 612/707 & 0.072 / 0.657 \\
Capacity-matched Fusion
& 84.27 & 0.289 & 0.000 & 547/707 & 0.512 / 0.286 \\
\textbf{RIVET}
& 83.33 & 0.281 & 0.000 & 521/707 & \textbf{0.651 / 0.205} \\
\bottomrule
\end{tabularx}
\end{table}

\section{Experiments}
\label{sec:experiments}

Our experiments examine both the predictive value of visual evidence and the destination of its decision effect. We first compare accuracy and ownership under shared evidence, then evaluate backbone reuse and task generalization, and finally examine the roles of evidence alignment and response control.

\noindent\textbf{Systems.}
We evaluate a Qwen2.5-VL-7B Base and four released reasoners: VLM-R$^3$-7B, Pixel-Reasoner-7B, DeepEyes-7B, and TreeVGR-7B~\citep{qwen25vl,vlmr3,pixelreasoner,deepeyes,treevgr}. With shared evidence, we compare prompting, pooled fusion, Unconstrained Residual Fusion (Unconstrained Fusion), calibrated aligned summation, and Capacity-matched Fusion. RIVET uses four frozen backbones: Qwen2.5-VL-7B, Qwen3-VL-8B, InternVL3.5-8B, and LLaVA-OneVision-2-8B~\citep{qwen25vl,qwen3vl,internvl35,llavaov2}.

\noindent\textbf{Metrics and protocol.}
CROSS-Bench accuracy uses native final answers or vector argmax, with equal root weights. Task gains use all 5,600 test roots; Table~\ref{tab:cross-main} uses the nested 960-root intervention matrix. $\Delta$Acc. denotes gains over Base in percentage points (pp). Posterior metrics use softmax-normalized option-identifier log likelihoods at the terminal answer slot or an interface's candidate vector; all required vectors must be available, complete, finite, and in the fixed candidate order. RIVET uses Ground unless noted; CV-Bench uses official source-balanced aggregation.

\subsection{Main Results: Accuracy and Ownership Diverge}

\noindent\textbf{Higher accuracy does not imply stronger ownership.}
Released reasoners benefit from clean evidence, yet remain sensitive to invalid evidence and transfer the clean effect incompletely (Table~\ref{tab:cross-main}a). With shared evidence, Unconstrained Fusion achieves the highest accuracy and clean utility, yet much of its effect remains with the original owner after valid rebinding. RIVET has lower matrix accuracy (83.33\% versus 84.58\%) but higher transfer (0.651 versus 0.072) and lower leakage (0.205 versus 0.657; Table~\ref{tab:cross-main}b). The ownership ordering also holds on common clean-active roots. Evidence can therefore be useful for prediction without directing its effect to the candidate supported by the current relation.

\noindent\textbf{Candidate conditioning matters beyond alignment.}
Calibrated Aligned Sum already achieves transfer of 0.603, showing the value of preserving evidence--candidate alignment. RIVET improves on this result while raising matrix accuracy from 79.79\% to 83.33\%. To examine the role of composition, Capacity-matched~Fusion retains the same evidence, training setup, output queries, and scalar control, but pools evidence independently of the queried candidate. RIVET raises transfer from 0.512 to 0.651 under this comparison, with the advantage persisting on common clean-active roots and across three paired fits. On common roots, RIVET also produces a larger positive new-owner increment before normalization. Candidate conditioning thus improves where the evidence acts, beyond supplying aligned inputs.

\noindent\textbf{Selective responses to semantic errors and evidence mismatch.}
Structural checks cannot reject a wrong reading whose schema is valid. On 400 clean/error pairs that hold the request and Base fixed, RIVET has lower mean posterior displacement than Capacity-matched Fusion (TV: 0.022 versus 0.058) under the same structural checks. Both methods retain positive clean-evidence gains.

\begin{samepage}
We next fix the request and vary whether the evidence concerns the requested entity and role. On 638 common roots under the new request, RIVET has similar mean matched log gain to Capacity-matched Fusion (0.229 versus 0.225), while reducing the gain from crossed evidence about the wrong entity or role (0.022 versus 0.063). The log-gain response thus favors relevant support while limiting the influence of mismatched evidence. The original request shows the same pattern; candidate-probability contrasts remain inconclusive.\par
\end{samepage}

\subsection{Backbone Reuse and Task Generalization}

\noindent\textbf{Task gains and ownership patterns persist across Base models.}
Ground improves all four backbones, averaging +5.70~pp on the 5,600 CROSS-Bench test roots, with transfer near 0.65 and leakage near 0.21 across models (Table~\ref{tab:generalization}a). Composer is trained jointly with Base posteriors from these four backbones, and the fitted interface is reused without backbone-specific refitting. Controlled adds a further 1.84~pp on average with the same Readers, Composer, and Bound, indicating remaining headroom in fact planning, source selection, and role assignment.

\begin{table}[!tb]
\caption{\textbf{Backbone reuse and task generalization.}
(a) CROSS-Bench accuracy on 5,600 roots (parentheses: gains over Base in pp); Ground ownership on the 960-root matrix with method-specific active sets.
(b) Frozen Ground task generalization averaged over four backbones with shared predicted evidence.}
\label{tab:generalization}
\centering
\footnotesize
\renewcommand{\arraystretch}{1.04}
\setlength{\tabcolsep}{3.2pt}

\textbf{(a) CROSS-Bench: backbone and route replication}\par
\vspace{2pt}

\begin{tabularx}{\linewidth}{
@{}
>{\raggedright\arraybackslash}X
>{\centering\arraybackslash}p{1.60cm}
*{2}{>{\centering\arraybackslash}p{2.70cm}}
>{\centering\arraybackslash}p{2.60cm}
@{}}
\toprule
Backbone
& Base\newline Acc.\ (\%)$\uparrow$
& Controlled\newline Acc.\ (\%)$\uparrow$
& Ground\newline Acc.\ (\%)$\uparrow$
& Ground $T_{\mathrm{auth}}\uparrow\,/\,L_{\mathrm{orig}}\downarrow$ \\
\midrule
Qwen2.5-VL-7B
& 66.34 & 73.18 (+6.84) & 71.30 (+4.96) & 0.651 / 0.205 \\
Qwen3-VL-8B
& 67.96 & 75.89 (+7.93) & 73.93 (+5.97) & 0.638 / 0.216 \\
InternVL3.5-8B
& 62.98 & 71.25 (+8.27) & 69.45 (+6.47) & 0.659 / 0.225 \\
LLaVA-OneVision-2-8B
& 66.00 & 73.11 (+7.11) & 71.39 (+5.39) & 0.647 / 0.196 \\
\midrule
\textbf{Mean over backbones}
& \textbf{65.82}
& \textbf{73.36 (+7.54)}
& \textbf{71.52 (+5.70)}
& \textbf{0.649 / 0.211} \\
\bottomrule
\end{tabularx}

\vspace{5pt}

\textbf{(b) Frozen Ground task generalization across external benchmarks}\par
\vspace{2pt}

\begin{tabular*}{\linewidth}{@{\extracolsep{\fill}}lcccc@{}}
\toprule
Endpoint / contrast
& V$^\star$Bench & MMStar & CV-Bench & MME-RW-L \\
\midrule
Mean Base Acc. (\%)
& 77.49 & 61.77 & 81.62 & 49.19 \\
Mean Capacity-matched Acc. (\%)
& 82.85 & 64.92 & 84.85 & 55.90 \\
Mean RIVET Acc. (\%)
& 83.38 & 64.72 & 85.14 & 56.23 \\
RIVET $-$ Base (pp)
& +5.89 & +2.95 & +3.52 & +7.04 \\
\bottomrule
\end{tabular*}
\end{table}

The task gains extend beyond CROSS-Bench. Frozen Ground improves accuracy over Base in all 16 backbone--benchmark pairs on V$^\star$Bench, MMStar, CV-Bench, and MME-RealWorld-Lite (MME-RW-L)~\citep{vstarbench,mmstar,cvbench,mmerealworld}, averaging +2.95 to +7.04~pp by benchmark (Table~\ref{tab:generalization}b). Capacity-matched Fusion also benefits from shared evidence, and the accuracy ranking between the two interfaces varies across tasks.

\begin{table}[!tb]
\caption{\textbf{State-only Oracle and architectural controls.}
(a) Oracle retains predicted bindings, active sets, and clean-effect normalizers per interface; owner labels remain hidden.
(b) Qwen2.5-VL-7B Ground: accuracy on 5,600 roots (parentheses: gains over Base in pp); relation metrics on the 960-root matrix, with ownership fixed to Full RIVET's 521 active roots and clean-effect normalizers.}
\label{tab:rivet-ablation}
\centering
\footnotesize
\renewcommand{\arraystretch}{1.18}
\setlength{\tabcolsep}{3.0pt}

\textbf{(a) State-only Oracle}\par
\vspace{2pt}

\begin{tabular*}{\linewidth}{@{\extracolsep{\fill}}lcc@{}}
\toprule
Decision interface
& Predicted state $T_{\mathrm{auth}}\uparrow/L_{\mathrm{orig}}\downarrow$
& Oracle state $T_{\mathrm{auth}}\uparrow/L_{\mathrm{orig}}\downarrow$ \\
\midrule
Unconstrained Fusion
& 0.072 / 0.657 & 0.101 / 0.612 \\
Capacity-matched Fusion
& 0.512 / 0.286 & 0.564 / 0.253 \\
\textbf{RIVET}
& \textbf{0.651} / \textbf{0.205} & \textbf{0.685} / \textbf{0.171} \\
\bottomrule
\end{tabular*}

\vspace{5pt}

\textbf{(b) Architectural controls}\par
\vspace{2pt}

\begin{tabularx}{\linewidth}{
@{}
>{\raggedright\arraybackslash}X
>{\centering\arraybackslash}p{2.65cm}
*{2}{>{\centering\arraybackslash}p{1.25cm}}
>{\centering\arraybackslash}p{2.35cm}
@{}}
\toprule
Variant
& Acc.\ (\%)$\uparrow$
& $G_{\mathrm{clean}}\uparrow$
& $A_{\mathrm{inv}}\downarrow$
& $T_{\mathrm{auth}}\uparrow\,/\,L_{\mathrm{orig}}\downarrow$ \\
\midrule
Qwen2.5-VL-7B (Base)
& 66.34 & 0.000 & 0.000 & -- \\
\midrule
\textbf{Full RIVET}
& 71.30 (+4.96) & 0.281 & \textbf{0.000} & 0.651 / 0.205 \\
\midrule
Hard program selection
& 70.52 (+4.18) & 0.179 & 0.004 & 0.083 / \textbf{0.056} \\
Shuffled evidence bindings
& 70.96 (+4.62) & 0.264 & 0.006 & 0.296 / 0.402 \\
Renormalize represented mass
& 70.75 (+4.41) & 0.221 & 0.014 & 0.421 / 0.344 \\
Pre-Compose evidence selection
& 70.63 (+4.29) & 0.234 & 0.007 & 0.507 / 0.276 \\
No scalar Bound ($D_1$)
& 70.98 (+4.64) & 0.265 & \textbf{0.000} & \textbf{0.676} / 0.194 \\
Candidate-wise Bound
& \textbf{71.55 (+5.21)} & \textbf{0.298} & \textbf{0.000} & 0.468 / 0.346 \\
\bottomrule
\end{tabularx}
\end{table}

\noindent\textbf{Independent relations show the same directional pattern.}
We further test transfer and selectivity on independently authored relations. On 320 attribute and spatial roots, RIVET has estimated transfer of 0.563 versus 0.447 for Capacity-matched Fusion. The ordering persists on common clean-active roots, although the paired comparison remains statistically inconclusive. A separate collection of 900 roots spans four relation types and tests selectivity under fixed requests. Across its 830 common scoreable roots, RIVET again preserves similar mean matched gain while reducing crossed gain at the request-supported candidate, consistent with the selectivity pattern on CROSS-Bench.

\subsection{Analysis of Evidence Alignment and Response Control}

\noindent\textbf{Better state estimates do not close the ownership gap.}
Replacing predicted visual states with Oracle states improves transfer and reduces leakage for both interfaces (Table~\ref{tab:rivet-ablation}a). Transfer rises from 0.512 to 0.564 for Capacity-matched Fusion and from 0.651 to 0.685 for RIVET, with evidence acquisition and bindings unchanged. Even with corrected states, the capacity-matched interface remains below RIVET with predicted states. More accurate readings help, but do not by themselves determine which candidate receives the evidence effect.

\noindent\textbf{Identity and complete support matter beyond accuracy.}
Shuffling evidence bindings retains most of RIVET's accuracy gain but more than halves transfer and increases leakage in frozen replay (Table~\ref{tab:rivet-ablation}b). Evidence can remain predictive even when its association with the supported candidate is weakened.

Reducing support before composition also harms ownership. Hard program selection weakens clean utility and nearly removes transfer despite low leakage; low leakage alone therefore does not establish successful rebinding. Renormalizing represented mass while removing unresolved-mass channels reduces transfer and increases leakage, as does pre-Compose evidence selection. Preserving bindings and uncertainty through composition matters beyond retaining enough information to answer correctly.

\begin{samepage}
\noindent\textbf{Bound calibrates an existing evidence response.}
Scalar Bound adds 0.32~pp in accuracy and improves clean utility over the unattenuated response $D_1$, while $D_1$ retains slightly better mean ownership scores (Table~\ref{tab:rivet-ablation}b). Refitting with candidate-wise control pushes accuracy higher still, but reduces transfer and increases leakage. Bound's benefit lies in calibrating the strength of an already composed response. Together, these comparisons support assigning candidate-specific evidence use to composition and leaving strength calibration to a shared scalar.\par
\end{samepage}

\section{Conclusion and Limitations}

CROSS-Bench shows that useful visual evidence need not direct its effect to the candidate supported by the underlying relation. RIVET addresses this gap through candidate-conditioned composition and separate strength control, improving transfer over Capacity-matched Fusion while retaining task gains over Base across frozen backbones and external benchmarks. These findings motivate evaluating evidence utility and effect destination together. The criterion focuses on positive contributions in finite candidate sets with explicit, annotatable relations; RIVET relies on registered typed state vocabularies and remains susceptible to schema-valid reading errors. Extending the analysis to contradictory or temporal evidence and open-ended generation, and relating decision-level effects to hidden-state causal mediation, are directions for future work.

\section*{Acknowledgments}
This study was supported by National Natural Science Foundation of China under Grants 61971404; in part by the Project of Technology and Engineering Center for Space Utilization, Chinese Academy of Sciences under Grant T503471; in part by the Youth Innovation Promotion Association of the Chinese Academy of Sciences under Grant 2019168.

\begingroup
\small
\setlength{\bibsep}{2pt plus 1pt}
\urlstyle{same}
\bibliography{references}
\bibliographystyle{plainnat}
\endgroup

\end{document}